\documentclass[9pt,conference]{IEEEtran}
\usepackage{amssymb,amsthm,amsmath,array}
\usepackage{graphicx}
\usepackage[caption=false,font=footnotesize]{subfig}
\usepackage{xspace}
\usepackage[sort&compress, numbers]{natbib}
\usepackage{stmaryrd}
\usepackage{xcolor}
\usepackage{mathtools}
\usepackage{float}
\usepackage{textcomp}
\usepackage{algorithm}
\usepackage{algorithmic}
\usepackage{multirow}
\usepackage{multicol}
\def\R{\mathbb{R}}  
\newcommand\bth[0]{{\boldsymbol{\theta}}} 

\newcommand{\bst}[0]{\boldsymbol{t}}

\newcommand\bsu[0]{{\boldsymbol{u}}}
\newcommand\bsv[0]{{\boldsymbol{v}}}

\begin{document}
\title{Mechanical Artifacts in Optical Projection Tomography: Classification and Automatic Calibration}
\author{\IEEEauthorblockN{
        Yan Liu\IEEEauthorrefmark{1}, 
        Jonathan Dong\IEEEauthorrefmark{1},
        Thanh-An Pham\IEEEauthorrefmark{1},
        Fran\c cois Marelli\IEEEauthorrefmark{2}\IEEEauthorrefmark{3}, and 
        Michael Unser\IEEEauthorrefmark{1}
    }
    \IEEEauthorblockA{
        \IEEEauthorrefmark{1}Biomedical Imaging Group, \'Ecole polytechnique f\'ed\'erale de Lausanne, Station 17, 1015 Lausanne, Switzerland\\
        \IEEEauthorrefmark{2}Idiap Research Institute, 1920 Martigny, Switzerland\\
        \IEEEauthorrefmark{3}\'Ecole polytechnique f\'ed\'erale de Lausanne, 1015 Lausanne, Switzerland}
}
\maketitle
\begin{abstract}
Optical projection tomography (OPT) is a powerful tool for biomedical studies. 
It achieves 3D visualization of mesoscopic biological samples with high spatial resolution using conventional tomographic-reconstruction algorithms. 
However, various artifacts degrade the quality of the reconstructed images due to experimental imperfections in the OPT instruments. 
While many efforts have been made to characterize and correct for these artifacts, they focus on one specific type of artifacts. This work has two contributions. First, we systematically document a catalog of mechanical artifacts based on a 3D description of the imaging system that uses a set of angular and translational parameters. 
Then, we introduce a calibration algorithm that recovers the unknown system parameters fed into the final 3D iterative reconstruction algorithm for a distortion-free volumetric image. 
Simulations with beads data and experimental results on a fluorescent textile fiber confirm that our algorithm successfully removes miscalibration artifacts in the reconstruction. 
\end{abstract}

\section{Introduction}
OPT is widely used in the 3D imaging of biological samples at mesoscopic scales in both brightfield and fluorescene configurations \cite{Walls2007, Walls2005, Wang:07, bassi:11, McGinty:11, Torres2021}. 
Analogous to X-ray computed tomography, a set of 2D projections are collected when the sample is rotated around an axis to different angles. 
A high-resolution 3D image of the inner structure of the sample is reconstructed using the filtered backprojection algorithm, assuming the imaging system is free of mechanical errors.
In practice, mechanical errors in the imaging system play a non-negligible role, especially in low-cost OPT systems.
The most common ones include an offset or tilt of the rotation axis which result in distortions of various degrees in the reconstruction due to a mismatch of the assumed imaging model and the actual physical process \cite{Walls2005, Zhang:20, Tang:16, Donath:06, Birk:10, Koskela19, vanderHorst:16, Ancora2017, Torres2021, Ramirez:19, Liu:22}. This work has been published in \cite{Liu:22}.

\section{Catalog of mechanical artifacts} 
\begin{figure}[ht]
\centering 
\includegraphics[width=0.7\columnwidth]{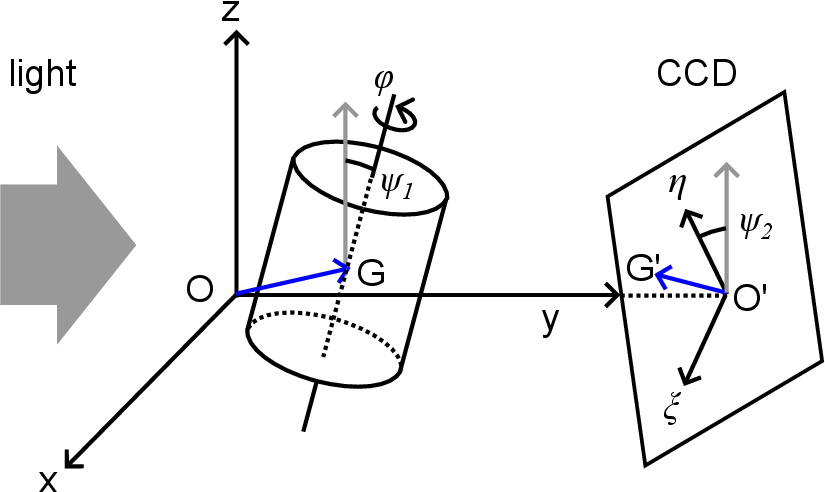}
\caption{
OPT imaging system with mechanical errors. 
(The lens system is omitted here for simplicity.)
The cylinder represents the tilted sample that rotates around its orientation axis.
The sample is placed in a 3D coordinate system $xyz$ with origin $O$.
The center of the cylinder is translated from $O$ to $G$. 
The detector plane is described by a 2D coordinate system $\xi\eta$ (with origin $O'$) perpendicular to the optical axis ($y$ axis). The projection of $O$ and $G$ in the detector plane are $O'$ and $G'$, respectively.
The angle $\psi_1$ represents the out-of-plane tilt angle between the orientation axis of the sample and the $z$-axis direction (gray vertical arrow).
The rotation angle $\psi_2$ represents the in-plane tilt angle of the detector plane around the optical axis.
}
\label{fig:geometry}
\end{figure}

\begin{figure}[ht] 
\centering
    \includegraphics[width=0.98\columnwidth]{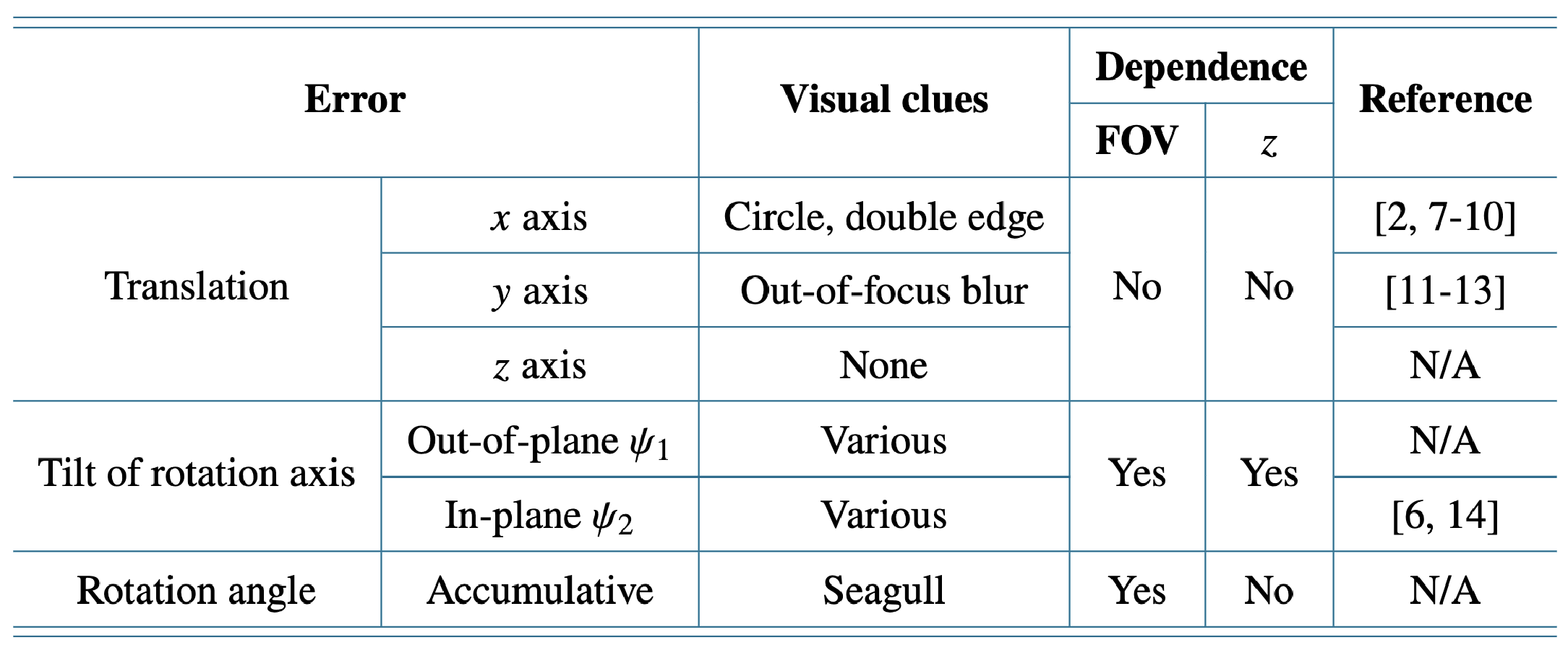}
\caption{Catalog of setup-related imaging artifacts and their dependence on the field of view (FOV) and on height (location along the $z$ axis).}
    \label{tab:dict}
\end{figure}

We introduce a catalog (See Fig. \ref{tab:dict}) of the imaging artifacts due to mechanical imperfections.
Based on a more precise description of the imaging geometry in Fig. \ref{fig:geometry}, we are able to document in a comprehensive fashion where the errors could occur in the imaging system, what visual clues are expected and their dependency on the field-of-view or the axial location, and existing works to address these issues.
This catalog helps experimentalists to identify artifacts they may encounter in the reconstructions, so as to correct for them either experimentally or computationally. 
\section{Automatic calibration of mechanical artifacts}
We propose a computational framework that automatically calibrates the parameters of our refined model of an OPT imaging system. 
This framework is able to detect the model mismatch between the simplistic forward model and the measurement data.
It outputs a set of calibrated system parameters and allows us to improve the reconstruction of the 3D volume.
We first formulate the forward model and the associated inverse problem. 
Then, we present our multiscale joint reconstruction-calibration algorithm and show how to remove the artifacts in both simulated and experimental data.
\subsection{Mathematical Model and Algorithm}
\begin{figure}[t]
    \centering
    \includegraphics[width=\columnwidth]{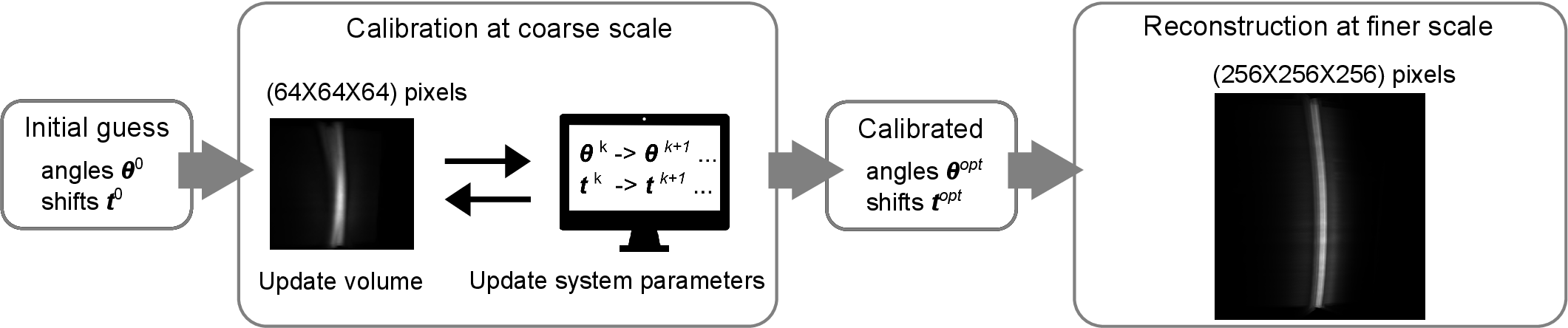}
    \caption{Workflow of the multiscale calibration-reconstruction algorithm.}
    \label{fig:flow-chart}
\end{figure}

To keep the forward model computationally efficient, we omit optical effects as is done in the common focal-sheet-scanning OPT setup \cite{Marelli:21}.
The forward model that describes straight-ray projections of a sample at any pose is as follows:
\begin{equation}\label{eq:our-continuous-forward}
    b^{\bth,\,\bst}(\bsv)=\mathcal{P}_{\bth}\{f\}(\bsv-\bst),
\end{equation}
where the compactly supported function $f(\bsu)\in L_2(\R^3)$, $\bsu=(x,y,z)$ represents the 3D sample to be reconstructed. 
$\mathcal{P}_{\bth}$ is the 3D X-ray transform. 
$b^{\bth,\,\bst}(\bsv)$ is the measurement in the detector plane for a location $\bsv=(\xi,\eta)$, a given sample orientation $\bth=(\boldsymbol{\varphi}, \psi_1, \psi_2) $, $\boldsymbol{\varphi}=(\varphi_1, \ldots, \varphi_P)$ is the set of rotation angles, and a shift vector $\bst=(t_1, t_2)$ in the detector plane.
To numerically implement the forward model, we sample Eq. \eqref{eq:our-continuous-forward} using the optimized Kaiser-Bessel window functions that are well-suited for tomographic imaging settings \cite{nilchian2015optimized} as basis and represent $f$ with a vector of coefficients $\mathbf{c}$:  
\begin{equation}\label{eq:final-discrete-forward}
    \mathbf{b}=\mathbf{H}(\boldsymbol{\theta}, \boldsymbol{t})\mathbf{c}.
\end{equation}
The corresponding inverse problem to recover the system parameters and reconstruct the coefficients that represents the 3D volume is formulated as:
\begin{equation}\label{eq:global-minimization}
    \mathbf{c}^*,(\boldsymbol{\theta}^*,\boldsymbol{t}^*)\in\arg\underset{\mathbf{c},\boldsymbol{\theta},\boldsymbol{t}}{\min}\left\{\frac{1}{2}\|\mathbf{H}(\boldsymbol{\theta},\boldsymbol{t})\mathbf{c}-\mathbf{b}\|^2_2\right\}.
\end{equation}
The proposed automatic calibration algorithm is detailed in Algorithm. \ref{algo:global-recon}.
In practice, we run our calibration algorithm on a downsampled version of the original measurement data.
This helps reduce the computational burden in the 3D setting, and avoid undesirable local minima.
The well-calibrated system parameters are then used to reconstruct an artifact-free 3D image by running an extra Step 3 in Algorithm \ref{algo:global-recon} at a finer scale. 
We summarize the workflow of the multiscale calibration-reconstruction algorithm in Fig. \ref{fig:flow-chart}. 
The reconstruction step of the algorithm is implemented using the GlobalBioIm library \cite{Soubies:19} and  
the calibration step uses functions from the Cryo-Em-Refinement library \cite{Zehni2020}.
\begin{algorithm}[t]
    \caption{Automatic Calibration}
    \label{algo:global-recon}
        \begin{algorithmic}[1]
            \REQUIRE $\boldsymbol{\theta}^0,\boldsymbol{t}^0,\mathbf{b},\lambda>0$
            \STATE $\boldsymbol{\theta}=\boldsymbol{\theta}^0, \boldsymbol{t}=\boldsymbol{t}^0$
            \WHILE{$\boldsymbol{\theta}$ and $\boldsymbol{t}$ not converged} 
                \STATE $\mathbf{c}\leftarrow\arg\underset{\mathbf{c}\in\R^N}{\min}\left\{\|\mathbf{H}(\boldsymbol{\theta},\boldsymbol{t})\mathbf{c}-\mathbf{b}\|^2_2\right\}$
                \STATE $(\boldsymbol{\theta},\boldsymbol{t})\leftarrow\arg\underset{\boldsymbol{\theta}\in\R^{P+2},\,\boldsymbol{t}\in\R^{2}}{\min}\left\{\|\mathbf{H}(\boldsymbol{\theta},\boldsymbol{t})\mathbf{c}-\mathbf{b}\|^2_2\right\}$
            \ENDWHILE    
        \RETURN $\boldsymbol{\theta}$ and $\boldsymbol{t}$
        \end{algorithmic}
\end{algorithm}
\begin{figure}[b]
    \centering
    \includegraphics[width=\columnwidth]{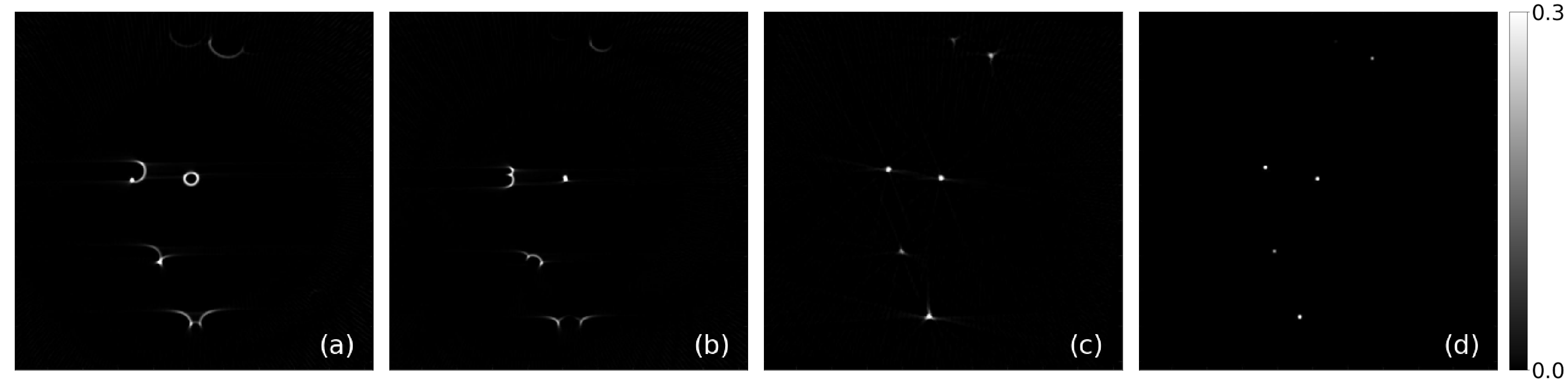}
    \caption{(a) Reconstruction with circle and seagull artifacts at one slice (147) without calibration. (b) Reconstruction result at slice 147 after a naive shift correction. (c) Calibrated reconstruction at slice 147. Both the circle and seagull artifacts are successfully removed. (d) Ground truth at slice 147. }
    \label{fig:simu_calibration}
\end{figure}
\subsection{Simulation and Experimental Results}
To validate Algorithm \ref{algo:global-recon}, we simulate an OPT system with random accumulative errors in the rotation angles and shifts in the rotation axis.  
We use the forward imaging model Eq. \ref{eq:final-discrete-forward} to generate projections.
In Fig. \ref{fig:simu_calibration}(a), we show that the reconstruction without any calibration or shift correction contains both the circle and seagull-shaped artifacts as expected. 
The reconstruction with only naive shift correction (Fig. \ref{fig:simu_calibration}(b)) still suffers from seagull artifacts due to residual model mismatch.
After applying our calibration algorithm, both the circle and the seagull artifacts are successfully removed all at once (Fig. \ref{fig:simu_calibration}(c)) and the reconstruction is very close to the ground-truth image in Fig.  \ref{fig:simu_calibration}(d).

We further validate our algorithm on an experimental dataset of a fluorescent textile fiber acquired with a lateral light-sheet illumination setup of which the rotation axis is tilted to four different values. 
We downsampled the measurements to a coarse scale of ($64 \times 64$)px per projection to speed up the computation. 
\begin{figure}[t]
    \centering
    \includegraphics[width=\columnwidth]{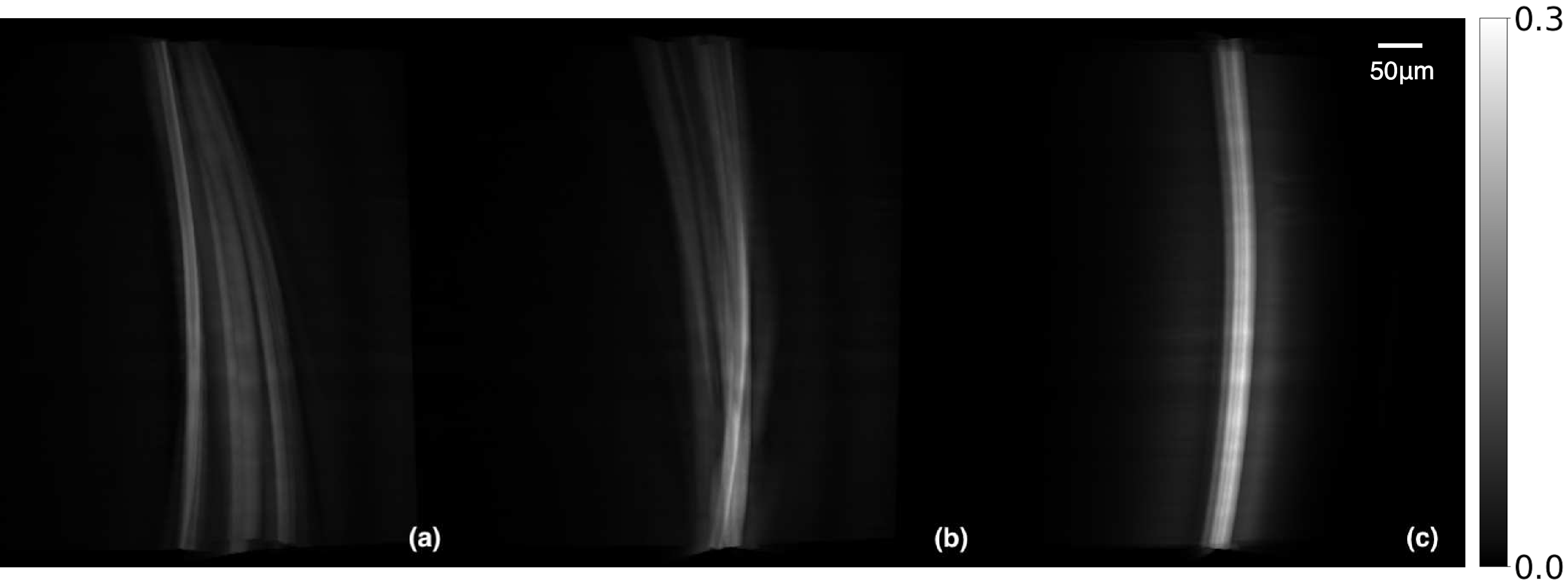}
    \caption{Textile fiber data. (a) - (b) Snapshots of the 3D visualization of the reconstruction results using the 2D FBP algorithm in a slice-by-slice fashion with no correction (a) and correcting only for the center of rotation (b). (c) 3D reconstruction result using the calibrated system parameters. }
    \label{fig:fiber}
\end{figure}
\begin{figure}[t]
    \centering
    \includegraphics[width=0.45\columnwidth]{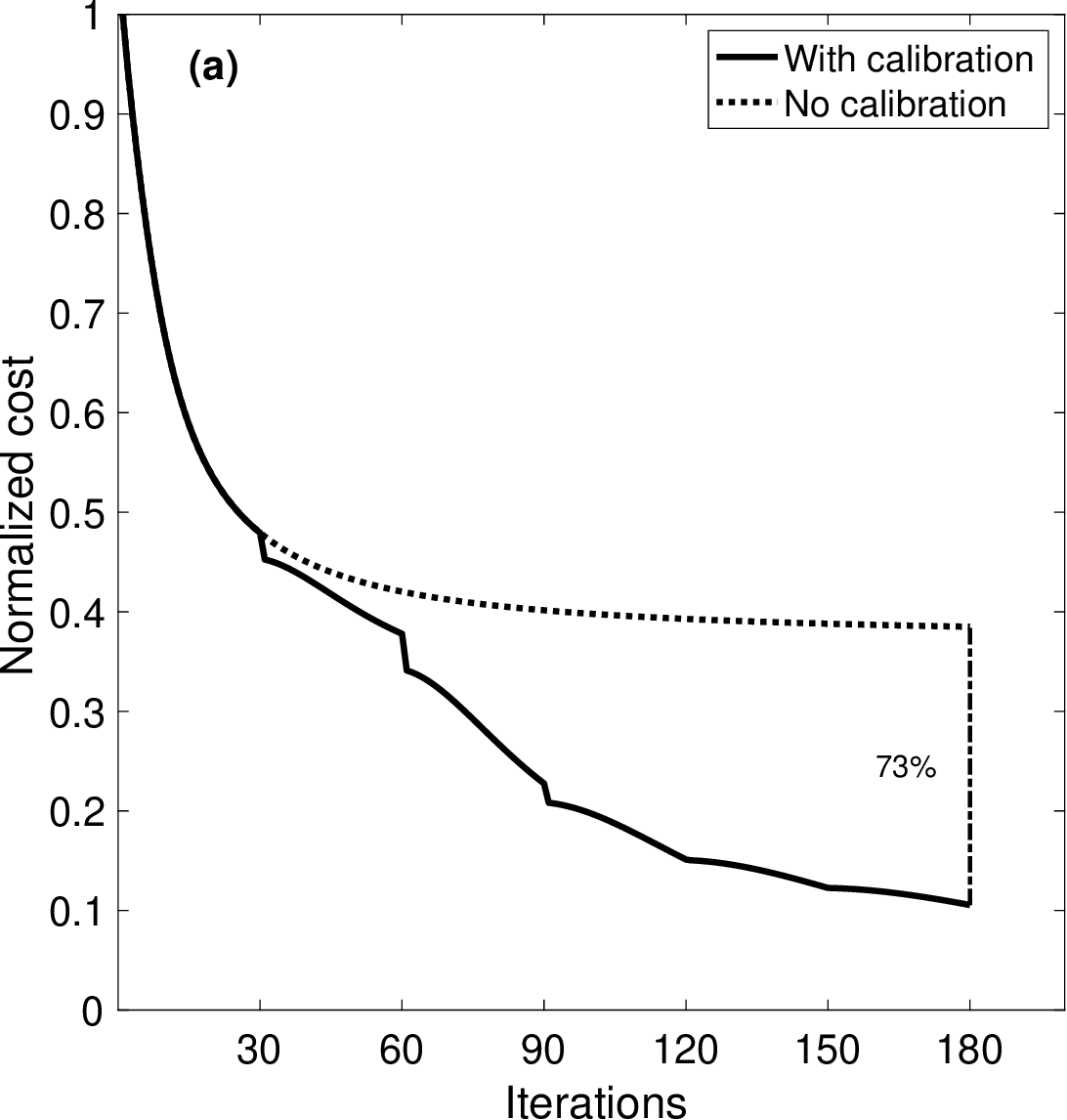}
    \includegraphics[width=0.45\columnwidth]{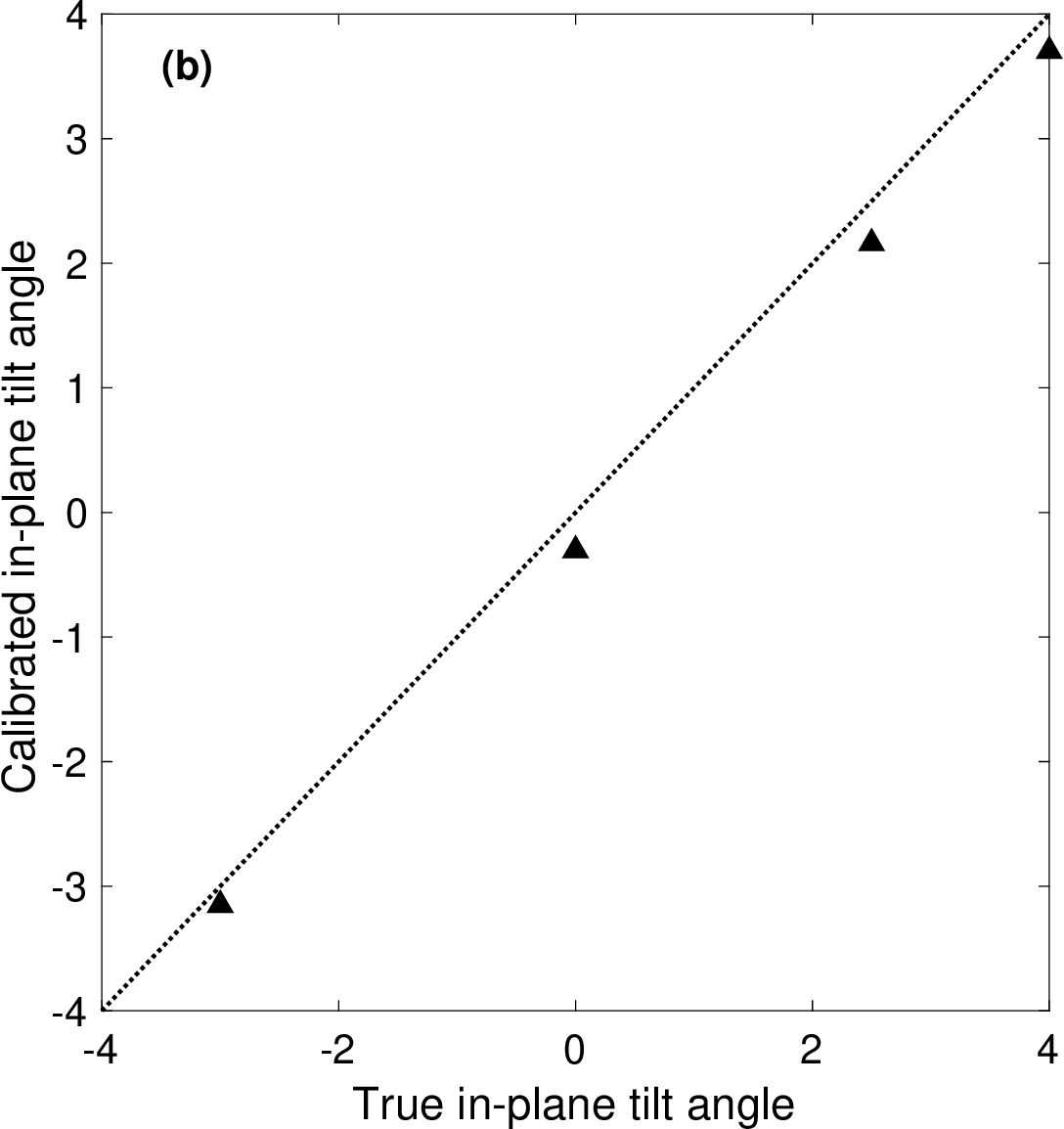}
    \caption{
    Convergence and accuracy. (a) Evolution of the cost function of the reconstruction calculated based on \eqref{eq:global-minimization} during the calibration process with (solid) and without (dashed) calibration. (b) The calibrated in-plane tilt angle against the true in-plane tilt angle for 4 different values indicated by the triangles. The dotted line represents $y=x$. The closer the triangles are to this line, the closer the calibrated angles are to the true angles.}
    \label{fig:stats}
\end{figure}
We show the four tilt angles after calibration in Fig. \ref{fig:stats}(b). 
In all four scenarios of different magnitudes and directions of tilt angles, the calibrated angle is very close to the controlled true value.
The 3D visualization of the reconstruction result of the fiber with a tilt angle of approximately 4 degrees is displayed in Fig. \ref{fig:fiber}. 
Without any correction, the reconstruction shows multiple ghost artifacts due to a combination of misaligned COR and tilt of the rotation axis in Fig. \ref{fig:fiber}(a). 
The severity of these ghost shadows are reduced after naive shift correction, as seen in Fig. \ref{fig:fiber}(b) but the top and middle parts of the reconstructed fiber still suffer.
The 3D reconstruction using the calibrated system parameters effectively removes all the shadows and achieves an artifact-free image.

More details on the code and data can be found in our paper \cite{Liu:22}.

\section{Conclusion}
We presented a comprehensive study of certain geometrical artifacts in a poorly calibrated OPT imaging system.
We described the 3D imaging geometry in detail. From there, we provided a systematic documentation of various mechanical artifacts, a more precise forward imaging model, and an automatic calibration algorithm that is able to recover the geometric parameters.
Reconstruction using the calibrated parameters produces clean artifact-free 3D images.
\bibliographystyle{IEEEtran}
\bibliography{references}

\begin{thebibliography}{10}
\providecommand{\url}[1]{#1}
\csname url@samestyle\endcsname
\providecommand{\newblock}{\relax}
\providecommand{\bibinfo}[2]{#2}
\providecommand{\BIBentrySTDinterwordspacing}{\spaceskip=0pt\relax}
\providecommand{\BIBentryALTinterwordstretchfactor}{4}
\providecommand{\BIBentryALTinterwordspacing}{\spaceskip=\fontdimen2\font plus
\BIBentryALTinterwordstretchfactor\fontdimen3\font minus
  \fontdimen4\font\relax}
\providecommand{\BIBforeignlanguage}[2]{{%
\expandafter\ifx\csname l@#1\endcsname\relax
\typeout{** WARNING: IEEEtran.bst: No hyphenation pattern has been}%
\typeout{** loaded for the language `#1'. Using the pattern for}%
\typeout{** the default language instead.}%
\else
\language=\csname l@#1\endcsname
\fi
#2}}
\providecommand{\BIBdecl}{\relax}
\BIBdecl

\bibitem{Walls2007}
\BIBentryALTinterwordspacing
J.~R. Walls, J.~G. Sled, J.~Sharpe, and R.~M. Henkelman, ``Resolution
  improvement in emission optical projection tomography,'' \emph{Physics in
  Medicine and Biology}, vol.~52, no.~10, pp. 2775--2790, apr 2007. [Online].
  Available: \url{https://doi.org/10.1088/0031-9155/52/10/010}
\BIBentrySTDinterwordspacing

\bibitem{Walls2005}
\BIBentryALTinterwordspacing
J.~R. Walls, J.~G. Sled, J.~Sharpe, and R.~Henkelman, ``Correction of artefacts
  in optical projection tomography,'' \emph{Physics in Medicine and Biology},
  vol.~50, no.~19, pp. 4645--4665, sep 2005. [Online]. Available:
  \url{https://doi.org/10.1088/0031-9155/50/19/015}
\BIBentrySTDinterwordspacing

\bibitem{Wang:07}
\BIBentryALTinterwordspacing
Y.~Wang and R.~K. Wang, ``Optimization of image-forming optics for transmission
  optical projection tomography,'' \emph{Appl. Opt.}, vol.~46, no.~27, pp.
  6815--6820, Sep 2007. [Online]. Available:
  \url{http://ao.osa.org/abstract.cfm?URI=ao-46-27-6815}
\BIBentrySTDinterwordspacing

\bibitem{bassi:11}
\BIBentryALTinterwordspacing
A.~Bassi, L.~Fieramonti, C.~D'Andrea, G.~Valentini, and M.~Mione, ``{In vivo
  label-free three-dimensional imaging of zebrafish vasculature with optical
  projection tomography},'' \emph{Journal of Biomedical Optics}, vol.~16,
  no.~10, pp. 1--4, 2011. [Online]. Available:
  \url{https://doi.org/10.1117/1.3640808}
\BIBentrySTDinterwordspacing

\bibitem{McGinty:11}
\BIBentryALTinterwordspacing
J.~McGinty, H.~B. Taylor, L.~Chen, L.~Bugeon, J.~R. Lamb, M.~J. Dallman, and
  P.~M.~W. French, ``In vivo fluorescence lifetime optical projection
  tomography,'' \emph{Biomed. Opt. Express}, vol.~2, no.~5, pp. 1340--1350, May
  2011. [Online]. Available:
  \url{http://opg.optica.org/boe/abstract.cfm?URI=boe-2-5-1340}
\BIBentrySTDinterwordspacing

\bibitem{Torres2021}
\BIBentryALTinterwordspacing
V.~C. Torres, C.~Li, W.~Zhou, J.~G. Brankov, and K.~M. Tichauer,
  ``Characterization of an angular domain fluorescence optical projection
  tomography system for mesoscopic lymph node imaging,'' \emph{Appl. Opt.},
  vol.~60, no.~1, pp. 135--146, Jan 2021. [Online]. Available:
  \url{http://opg.optica.org/ao/abstract.cfm?URI=ao-60-1-135}
\BIBentrySTDinterwordspacing

\bibitem{Zhang:20}
\BIBentryALTinterwordspacing
H.~Zhang, L.~Waldmann, R.~Manuel, H.~Boije, T.~Haitina, and A.~Allalou,
  ``{zOPT}: an open source optical projection tomography system and methods for
  rapid 3{D} zebrafish imaging,'' \emph{Biomed. Opt. Express}, vol.~11, no.~8,
  pp. 4290--4305, Aug 2020. [Online]. Available:
  \url{http://www.osapublishing.org/boe/abstract.cfm?URI=boe-11-8-4290}
\BIBentrySTDinterwordspacing

\bibitem{Tang:16}
X.~Tang, M.~van't Hoff, J.~Hoogenboom, Y.~Guo, F.~Cai, G.~Lamers, and
  F.~Verbeek, ``Fluorescence and bright-field {3D} image fusion based on
  sinogram unification for optical projection tomography,'' in \emph{2016 IEEE
  International Conference on Bioinformatics and Biomedicine (BIBM)}, 2016, pp.
  403--410.

\bibitem{Donath:06}
\BIBentryALTinterwordspacing
T.~Donath, F.~Beckmann, and A.~Schreyer, ``Automated determination of the
  center of rotation in tomography data,'' \emph{J. Opt. Soc. Am. A}, vol.~23,
  no.~5, pp. 1048--1057, May 2006. [Online]. Available:
  \url{http://opg.optica.org/josaa/abstract.cfm?URI=josaa-23-5-1048}
\BIBentrySTDinterwordspacing

\bibitem{Birk:10}
\BIBentryALTinterwordspacing
U.~J. Birk, M.~Rieckher, N.~Konstantinides, A.~Darrell, A.~Sarasa-Renedo,
  H.~Meyer, N.~Tavernarakis, and J.~Ripoll, ``Correction for specimen movement
  and rotation errors for in-vivo optical projection tomography,''
  \emph{Biomed. Opt. Express}, vol.~1, no.~1, pp. 87--96, Aug 2010. [Online].
  Available: \url{http://opg.optica.org/boe/abstract.cfm?URI=boe-1-1-87}
\BIBentrySTDinterwordspacing

\bibitem{Koskela19}
O.~Koskela, T.~Montonen, B.~Belay, E.~Figueiras, S.~Pursiainen, and
  J.~Hyttinen, ``Gaussian light model in brightfield optical projection
  tomography,'' \emph{Scientific Reports}, vol.~9, 09 2019.

\bibitem{vanderHorst:16}
\BIBentryALTinterwordspacing
J.~van~der Horst and J.~Kalkman, ``Image resolution and deconvolution in
  optical tomography,'' \emph{Opt. Express}, vol.~24, no.~21, pp.
  24\,460--24\,472, Oct 2016. [Online]. Available:
  \url{http://opg.optica.org/oe/abstract.cfm?URI=oe-24-21-24460}
\BIBentrySTDinterwordspacing

\bibitem{Ancora2017}
\BIBentryALTinterwordspacing
D.~Ancora, D.~D. Battista, G.~Giasafaki, S.~Psycharakis, E.~Liapis,
  A.~Zacharopoulos, and G.~Zacharakis, ``{Optical projection tomography via
  phase retrieval algorithms for hidden three dimensional imaging},'' in
  \emph{Quantitative Phase Imaging III}, vol. 10074.\hskip 1em plus 0.5em minus
  0.4em\relax SPIE, 2017, p. 100741E. [Online]. Available:
  \url{https://doi.org/10.1117/12.2252894}
\BIBentrySTDinterwordspacing

\bibitem{Ramirez:19}
\BIBentryALTinterwordspacing
P.~P. Vallejo~Ramirez, J.~Zammit, O.~Vanderpoorten, F.~Riche, F.-X. Bl{\'e},
  X.-H. Zhou, B.~Spiridon, C.~Valentine, S.~E. Spasov, P.~W. Oluwasanya,
  G.~Goodfellow, M.~J. Fantham, O.~Siddiqui, F.~Alimagham, M.~Robbins,
  A.~Stretton, D.~Simatos, O.~Hadeler, E.~J. Rees, F.~Str{\"o}hl, R.~F. Laine,
  and C.~F. Kaminski, ``{OptiJ}: Open-source optical projection tomography of
  large organ samples,'' \emph{Scientific Reports}, vol.~9, no.~1, p. 15693,
  Oct 2019. [Online]. Available:
  \url{https://doi.org/10.1038/s41598-019-52065-0}
\BIBentrySTDinterwordspacing

\bibitem{Liu:22}
\BIBentryALTinterwordspacing
Y.~Liu, J.~Dong, T.~an~Pham, F.~Marelli, and M.~Unser, ``Mechanical artifacts
  in optical projection tomography: classification and automatic calibration,''
  \emph{Opt. Continuum}, vol.~1, no.~12, pp. 2577--2589, Dec 2022. [Online].
  Available:
  \url{https://opg.optica.org/optcon/abstract.cfm?URI=optcon-1-12-2577}
\BIBentrySTDinterwordspacing

\bibitem{Marelli:21}
F.~Marelli and M.~Liebling, ``Optics versus computation: Influence of
  illumination and reconstruction model accuracy in focal-plane-scanning
  optical projection tomography,'' in \emph{2021 IEEE 18th International
  Symposium on Biomedical Imaging (ISBI)}, 2021, pp. 567--570.

\bibitem{nilchian2015optimized}
M.~Nilchian, J.~P. Ward, C.~Vonesch, and M.~A. Unser, ``Optimized
  kaiser–bessel window functions for computed tomography,'' \emph{IEEE
  Transactions on Image Processing}, vol.~24, pp. 3826--3833, 2015.

\bibitem{Soubies:19}
\BIBentryALTinterwordspacing
E.~Soubies, F.~Soulez, M.~T. McCann, {\exhyphenpenalty9999\relax{}T.-a.}~Pham,
  L.~Donati, T.~Debarre, D.~Sage, and M.~Unser, ``Pocket guide to solve inverse
  problems with {GlobalBioIm},'' \emph{Inverse Problems}, vol.~35, no.~10, p.
  104006, sep 2019. [Online]. Available:
  \url{https://doi.org/10.1088/1361-6420/ab2ae9}
\BIBentrySTDinterwordspacing

\bibitem{Zehni2020}
M.~Zehni, L.~Donati, E.~Soubies, Z.~Zhao, and M.~Unser, ``Joint angular
  refinement and reconstruction for single-particle cryo-em,'' \emph{IEEE
  Transactions on Image Processing}, vol.~29, pp. 6151--6163, 2020.

\end{thebibliography}
\end{document}